\documentclass[conference,comsoc]{IEEEtran}
\usepackage{cite}

\ifCLASSINFOpdf
  \usepackage[pdftex]{graphicx}
\else
  \usepackage[dvips]{graphicx}
\fi
\usepackage[cmex10]{amsmath}
\usepackage{algorithmic}
\usepackage{array}
\ifCLASSOPTIONcompsoc
  \usepackage[caption=false,font=normalsize,labelfont=sf,textfont=sf]{subfig}
\else
  \usepackage[caption=false,font=footnotesize]{subfig}
\fi
\usepackage{dblfloatfix}
\usepackage{booktabs}
\usepackage{url}
\usepackage{float}
\usepackage{multirow}
\usepackage{enumerate}
\usepackage{amsthm}
\usepackage{nidanfloat}
\usepackage{threeparttable}
\usepackage{svg}

\newtheoremstyle{mystyle}
  {}
  {}
  {\itshape}
  {}
  {\bfseries}
  {.}
  { }
  {}
\theoremstyle{mystyle}

\newenvironment{talign*}
 {\let\displaystyle\textstyle\csname align*\endcsname}
 {\endalign}
\usepackage{arydshln}
\usepackage[normalem]{ulem}
\usepackage{hyperref}
\usepackage{amsmath}
\DeclareMathOperator*{\argmax}{argmax} 
\usepackage{xurl}

\begin{document}

%
\title{NoiseCAM: Explainable AI for the Boundary Between Noise and Adversarial Attacks
}

%

\author{%
    Wenkai Tan$^{~1a}$, Justus Renkhoff$^{~1a}$, Alvaro Velasquez$^{2b}$, Ziyu Wang$^{5g}$, Lusi Li$^{5h}$, Jian Wang$^{3d}$\\ Shuteng Niu$^{4e}$, Fan Yang $^{1c}$, Yongxin Liu $^{1c}$, Houbing Song$^{6}$\\
    $^{1}$Embry-Riddle Aeronautical University, FL 32114 USA,
    $^{2}$University of Colorado Boulder, CO 80309 USA\\
    $^{3}$University of Tennessee at Martin, TN 38237 USA,
    $^{4}$Bowling Green State University, OH 43403 USA, \\
    $^{5}$Old Dominion University, VA 23529 USA,
    $^{6}$University of Maryland, Baltimore County, Baltimore, MD 21250 USA\\

    $^{a}$\{tanw1, renkhofj\}@my.erau.edu,
    $^{b}$alvaro.velasquez@colorado.edu,
    $^{c}$\{LIUY11, YANGF1\}@erau.edu,\\
    $^{d}$jwang186@utm.edu, $^{e}$sniu@bgsu.edu
    $^{f}$songh@umbc.edu,
    $^{g}$zwang007@odu.edu, $^{h}$lusili@cs.odu.edu\\
    \thanks{}
}

\markboth{IEEE Internet of Things Journal,~Vol.~11, No.~4, May~2021}%
{Shell \MakeLowercase{\textit{et al.}}: Bare Demo of IEEEtran.cls for Journals}
\IEEEtitleabstractindextext{%
\begin{abstract}
Deep Learning (DL) and Deep Neural Networks (DNNs) are widely used in various domains. However, adversarial attacks can easily mislead a neural network and lead to wrong decisions. Defense mechanisms are highly preferred in safety-critical applications. In this paper, firstly, we use the gradient class activation map (GradCAM) to analyze the behavior deviation of the VGG-16 network when its inputs are mixed with adversarial perturbation or Gaussian noise. In particular, our method can locate vulnerable layers that are sensitive to adversarial perturbation and Gaussian noise. We also show that the behavior deviation of vulnerable layers can be used to detect adversarial examples. Secondly, we propose a novel NoiseCAM algorithm that integrates information from globally and pixel-level weighted class activation maps. Our algorithm is highly sensitive to adversarial perturbations and will not respond to Gaussian random noise mixed in the inputs. Third, we compare detecting adversarial examples using both behavior deviation and NoiseCAM, and we show that NoiseCAM outperforms behavior deviation modeling in its overall performance. Our work could provide a useful tool to defend against certain types of adversarial attacks on deep neural networks.

\end{abstract}

}

\IEEEoverridecommandlockouts
\maketitle
\IEEEdisplaynontitleabstractindextext
\IEEEpeerreviewmaketitle

\section{Introduction}

Artificial intelligence and deep learning have great potential to tackle a wide range of engineering and scientific challenges. For example, Deep Learning have been widely applied in medical imaging to reduce delay and human efforts in disease diagnoses \cite{niu2021distant}. In addition, deep neural networks are extremely good at dealing with complicated scenarios, making DL a promising tool for safety and security domains where complicated patterns are difficult to define explicitly \cite{liu2021machine,wang2021counter}.

However, the robustness and reliability of Deep Neural Networks (DNNs) have raised concerns among researchers. Studies have shown that these networks, such as the VGG-16 model \cite{simonyan2014very}, can be fooled by manipulated images that are undetectable to the human eye \cite{10.1145/3132747.3132785,miller2020adversarial,luo2018towards}. In such cases, the altered pixels possess pseudo-random properties, leading to doubts about the reliability and confidence of DNNs in environments with natural Gaussian noise \cite{englert2019machine,ignatiev2020towards}.

For mitigation, on the one hand, some solutions are proposed to increase the robustness of DNNs by augmenting their training with perturbed samples or introducing a robust loss term \cite{wu2021wider,gong2021maxup,gao2019convergence}. These approaches encourage DNNs to treat a slightly perturbed image as its origin. In this context, finding and incrementally training DNNs with adversarial examples is analogous to fuzzy testing. Some representative frameworks have been proposed, such as DLFuzz \cite{9099600}, DeepXplore \cite{10.1145/3132747.3132785}, DeepHunter \cite{xie2019deephunter}, and TensorFuzz \cite{odena2019tensorfuzz}. In particular, in one of our previous works of a white-box fuzz testing framework, DLFuzz, we derive adversarial perturbation by misleading the neural classifier and maximizing the neuron coverage simultaneously.

One common feature of these adversarial example-enabled neural network fuzzy testing frameworks is that they not only discover adversarial examples but also try to maximize the activation rates of neurons, also known as neuron coverage \cite{yang2022revisiting, harel2020neuron}. The neuron coverage ratio describes how many neurons are activated during a prediction. DLFuzz \cite{9099600} adapts this concept from DeepXplore \cite{10.1145/3132747.3132785} and tries to optimize this metric by generating adversarial examples and maximizing the prediction difference between the original and the adversarial images. Higher neuron coverage usually contributes positively to the robustness of DNNs.

However, training DNNs on perturbed samples incrementally is computationally expensive and can reduce classification accuracy \cite{lamb2019interpolated}. Moreover, it is difficult to find a balance between accuracy and adversarial robustness. Defending existing DNNs against adversarial examples without specifically retraining them is preferred. Defense-GAN \cite{samangouei2018defense} trains a defensive generative adversary network (GAN) on natural inputs. A noticeable behavior deviation can be detected when adversarial examples are fed into the defensive GAN. Instead of modeling the inputs directly, I-Defender \cite{zheng2018robust} models the output distributions of fully connected hidden layers for each class. Then it uses statistical testing to reject adversarial examples. Adversarial perturbations can be treated as additive noise. Therefore, similar approaches, such as denoising autoencoders, can be used to purify the input of DNNs \cite{gondara2016medical, gu2014towards, yadav2022integrated,hwang2019puvae}.

Most current efforts view DNNs as black-box models and fail to analyze adversarial attacks in an explainable manner. In our prior work \cite{9894322}, we used ImageNet database \cite{5206848} and then manipulate them with DLFuzz \cite{9099600} to generate white-box  adversarial attacks examples. We also used Grad-CAM \cite{Selvaraju_2017_ICCV} enabled class activation maps in VGG-16 \cite{simonyan2014very}. We compare the model's responses to adversarial and statistically similar Gaussian noise mixed images, revealing the potential of class activation maps for detecting adversarial examples. This paper examines the potential of modeling behavior deviation in vulnerable VGG-16 layers to identify adversarial examples and introduces NoiseCAM, a novel method that integrates GradCAM++ \cite{chattopadhay2018grad} and LayerCAM \cite{9462463} to detect adversarial examples in an interpretable manner. NoiseCAM is more effective than our behavior modeling method in detecting adversarial examples. The contributions of our work are as follows.
\begin{itemize}
    \item We propose an interpretable method for detecting compromised layers in a neural network under adversarial attacks by modeling behavior deviation. Our findings show that the deeper layers in VGG-16 are more sensitive to adversarial perturbations and Gaussian random noise.
    \item We investigate the use of behavior deviation modeling in vulnerable convolution layers under non-adversarial scenarios to obtain decision thresholds for detecting adversarial examples, but found it to be unreliable.
    \item We propose NoiseCAM, a new algorithm for detecting adversarial attacks on DNNs in an interpretable manner. Our experiments demonstrate that NoiseCAM is highly sensitive to adversarial noise while being impervious to Gaussian random noise, even when they have statistically similar properties.
\end{itemize}

The remainder of this paper is organized as follows: A literature review of related work is presented in Section~\ref{sectRW}. We present the methodology in Section~\ref{sectMM}. Evaluation and discussion are presented in Section~\ref{sectEED} and conclusions in Section~\ref{sectCC}.

\section{Related Work}

\label{sectRW}


\textit{Fuzzy Testing Enabled Adversarial Example Generation on DNNs}: Given a network model, an adversarial example is generated by introducing small and imperceptible perturbations to a given seed image to cause misclassifications. There are several procedures, such as FGSM \cite{goodfellow2014explaining}, to generate adversarial examples. In general, this is a white-box fuzzy testing procedure, one has to know the parameters of the target neural classifier and then solve the optimization problem so that the perturbations should maximize the classification loss and minimize the difference between the perturbed and original image. Defending DNNs against adversarial examples can increase the robustness of the model.
\begin{figure}[t]
	\centering
	\includegraphics[width=0.85\linewidth]{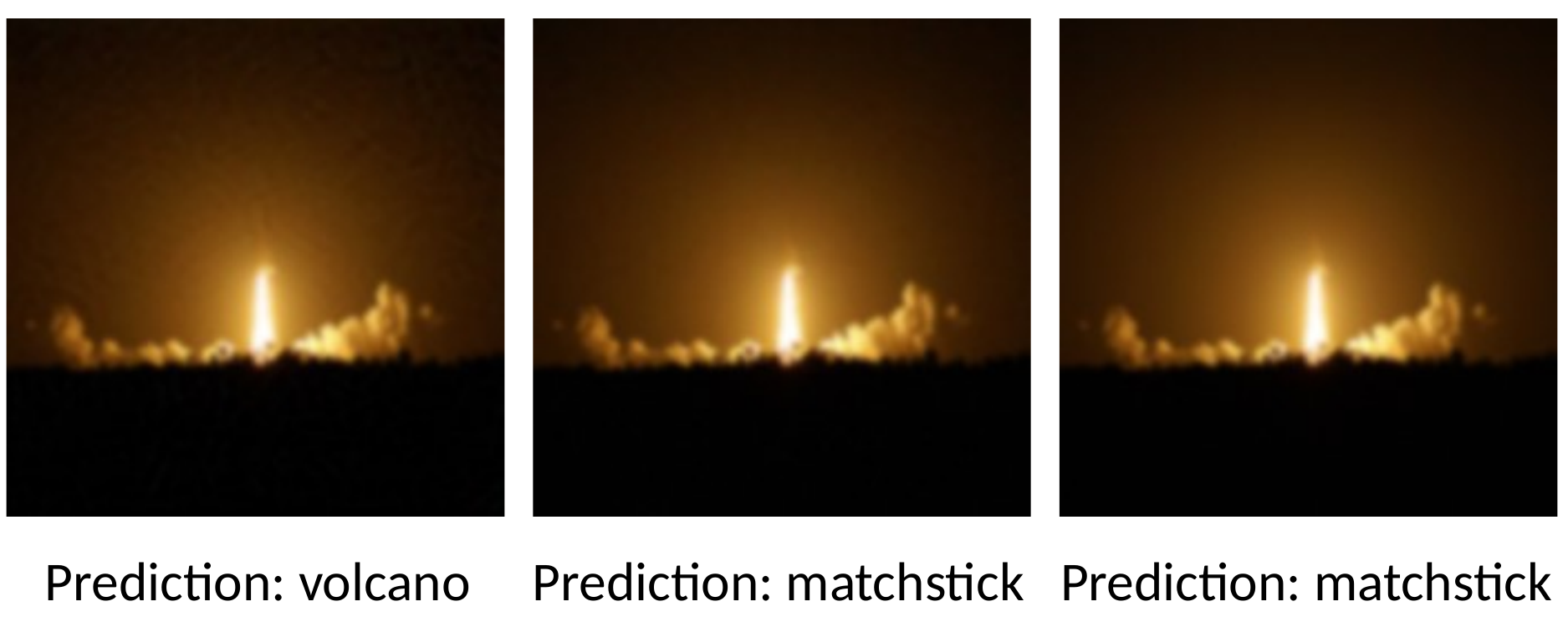}
	\caption{The effect of an adversarial  noise removal filter on VGG-16 model classification accuracy. We apply a Gaussian Blur filter with radius 1.5 to three distinct scenarios: a) an adversarial input, b) an input with Gaussian noise, and c) the original input (labeled as Space Shuttle). All three inputs subjected to the Gaussian Blur filter are incorrectly predicted by the VGG-16 network.
	}
	\label{figGaussBlur}
\end{figure}

\textit{Explainalbe AI (XAI) for Visual Explanations of DNNs}: Explainable AI (XAI) methods such as Gradient-weighted Class Activation Maps (Grad-CAMs) \cite{selvaraju2017grad} and Local Interpretable Model-Agnostic Explanations (LIME) \cite{ribeiro2016should} aim to improve the interpretability of DNNs. They have the ability to explain the prediction of black-box models and can therefore improve their trustworthiness. Grad-CAM calculates a heatmap highlighting different areas of an image in different colors. These colors visualize how much an area positively contributes to a certain prediction. LIME highlights pixels that contribute positively or negatively based on a threshold. These pixels can be represented in different colors, as in Grad-CAM. With Grad-CAM, it is possible to identify not only which areas contribute positively and negatively to a certain prediction, but also how much an area influences a decision \cite{cian2020evaluating}. Numerous efforts use these methods to increase confidence in machine learning models \cite{2008.02312} \cite{poerner-etal-2018-evaluating} \cite{app112110417}.

\textit{Countermeasures for Adversarial Attacks}: The defense against adversarial attacks has been extensively studied with several methods summarized in \cite{9013065}. The traditional method to defend test-time evasion (TTE) attacks is to remove all potential perturbations for the input images, representative methods are: Principal Component Analysis (PCA), blur filters, and autoencoders. However, these methods can also potentially decrease the accuracy of the DNN classifier. 

For example, in Figure~\ref{figGaussBlur}, we applied a Gaussian Blur filter with radius 1.5 to three different scenarios to protect the VGG-16 model against adversarial examples: a) an adversarial input, b) input with Gaussian noise and c) the original input. All three inputs with Gaussian Blur filter are erroneously predicted by the VGG-16 network. One reason is that DNN classifiers compress the input image into a relatively low resolution; e.g., VGG-16 compresses the input into 224x224. When we apply a blur filter or PCA to input images, the loss of information will lead to incorrect predictions. Although traditional methods can easily eliminate the perturbation, it also affect the classification results.

\section{Methodology}
\label{sectMM}
\subsection{Problem Definition}
\begin{figure}[htbp]
    \centering
    \includegraphics[width=0.9\linewidth]{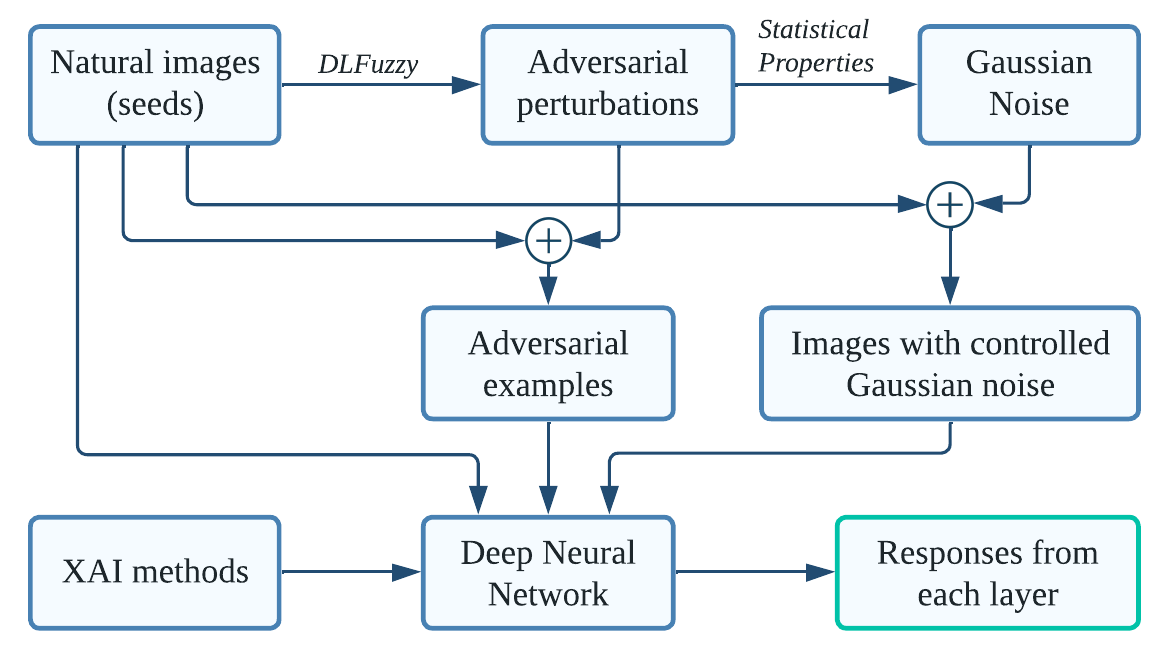}
    \caption{Data pre-processing and analysis procedure \cite{9894322}}
    \label{figWorkFlow}
\end{figure}

In order to safeguard DNNs against adversarial attacks, it is crucial to understand how these examples lead to misclassifications. To achieve this, we utilize XAI techniques to analyze the classification decision process of the VGG-16 model, layer by layer. Figure~\ref{figWorkFlow} provides a concise overview of our investigation method. 

\subsection{Data Preparation}
We used randomly selected images as seeds from the ImageNet validation dataset, and we derive their adversarial perturbations, or adversarial noise, denoted as:
\begin{align}
\label{eqADVPerturb}
    \mathbf{N}_a = \mathrm{ADV}(\mathbf{S}, \varepsilon)
\end{align}
where $\mathrm{ADV}(\cdot)$ is the adversarial perturbation function, $\varepsilon$ is the strength of the perturbation, and $\mathbf{S}$ denotes a seed image.  In this work, DLFuzz is configured to generate adversarial noise and maximize neuron coverage simultaneously. We generate the same amount of Gaussian noise $\mathbf{N}_g$, defined as random noise, with statistical properties similar to adversarial perturbations. Consequently, we use the expected value, standard deviation and shape of $\mathbf{N}_a$ to generate $\mathbf{N}_g$ as:
\begin{align}
\label{eqGaussianNoise}
    \mathbf{N}_g \sim \mathcal{N}(\mu_a, \Sigma_a)    \,
\end{align}
where $\mathcal{N}(\cdot)$ denotes a normal distribution, $\mu_a$ denotes the expected value, and $\Sigma_a$ denotes the standard deviation. We let $\mu_a = \overline{\boldsymbol{N}_a}$ and $\Sigma_a = \sigma(\boldsymbol{N}_a)$. Based on this distribution, we generate random noise $\mathbf{N}_g$ represented as a matrix with the same shape as $\mathbf{N}_a$. 
In this way, the augmented input $\mathbf{I}$ to the DNN becomes:
\begin{align}
    \mathbf{I} = \{\mathbf{S}, \mathbf{S}+\mathbf{N}_a, \mathbf{S}+\mathbf{N}_g\}
\end{align}
where $\mathbf{S}+\mathbf{N}_a$ and $\mathbf{S}+\mathbf{N}_g$ are adversarial examples and the noisy version of the seed image. 


\subsection{Adversarial Example Generation} 
\label{sec:DLFuzz}
We use DLFuzz \cite{9099600} to calculate perturbations and apply it to images from the ImageNet \cite{ILSVRC15} dataset. Mathematically, we are conducting a white-box attack on the neural network. The procedure of deriving additive perturbations is regarded as an optimization problem:
\begin{align}
    \argmax_{\mathbf{N}_a,~||\mathbf{N}_a|| \leq \delta}\left[ \sum^K_{i=1}c_i-c +\lambda\sum^m_{i=0}n_i\right]
\end{align}
where $\delta$ restricts the magnitude of the adversarial perturbation, $c$ is the prediction on a given seed image, $c_i$ is one of the top $K$ candidate predictions, $n_i$ is one of the activation values in the $m$ selected neurons, and $\lambda$ is a constant to balance between activating more neurons and obtaining adversarial examples.

DLFuzz manipulates the perturbations to mislead a tested network to make wrong predictions and simultaneously activate the selected neurons. DLFuzz ensures that adversarial examples are generated more efficiently during a test by maximizing neuron coverage. This is based on the assumption that this will trigger more logic in the network and thus provoke and detect more erroneous behaviors \cite{10.1145/3132747.3132785}. 

After obtaining an effective adversarial perturbation for each seed, we amplify the perturbations with a total of five different ratios, defined as the perturbation strength. These are 25\%, 50\%, 100\%, 200\% and 400\%.

The additive perturbation is then added to the original image to form adversarial examples. Unlike the Gaussian random noise that is evenly distributed on the image, adversarial perturbations are usually presented clustered with a certain pattern. These adversarial perturbations can lead to the misactivation of the wrong convolutional filters and can even be visualized on class activation maps.

\subsection{Network Behavior Deviation Modeling}
\label{sectBDD}
For a given convolution layer $l$, its response to an input image, e.g., a seed image, can be visualized using its grad-CAM heatmap, defined as:
\begin{align}
    \mathbf{H(\mathbf{S})} = R \left[ sign \left( \dfrac{\mathbf{J(\mathbf{S})}}{\partial\mathbf{\theta_l}} \right) \right]
\end{align}
where $\mathbf{J(\mathbf{S})}$ denotes the classification score of a selected category $\mathbf{S}$ and $\mathbf{\theta_l}$ denotes the parameters of layer $l$. The function $R(\cdot)$ denotes the operation that reshapes and interpolates the derived gradients to the same dimension and size as $\mathbf{S}$. The derivative values display the focal areas of $l$ in $\mathbf{S}$. Our hypothesis is that the focal regions under attack in the neural network could differ under adversarial and natural images. Moreover, we use cosine similarity to calculate the degree of behavior deviation as:
\begin{align}
     \label{eqCosineSim}
     \mathbf{G}[\mathbf{S_1},\mathbf{S_2}] = \dfrac{vec(\mathbf{H(\mathbf{S_1}))} \cdot vec(\mathbf{H(\mathbf{S_2}))}}{\| vec(\mathbf{H(\mathbf{S_1}))} \|\cdot \|  vec(\mathbf{H(\mathbf{S_2}))} \|}
    \end{align}
where $\mathbf{H(S_1)}$ and $\mathbf{H(S_2)}$ denote Grad-CAM heatmaps of different images. These are vectorized and normalized for similarity calculation. For a specific layer and a seed image $\mathbf{S}$, we calculate and compare two types of behavior deviation.
\begin{align}
    \mathbf{D}_a = \mathbf{G}[\mathbf{S}, \mathbf{S}+\mathbf{N}_a]\\
    \mathbf{D}_g =\mathbf{G}[\mathbf{S}, \mathbf{S}+\mathbf{N}_g]
\end{align}

\begin{figure*}[t]
    \centering
    \includegraphics[width = 0.85\textwidth]{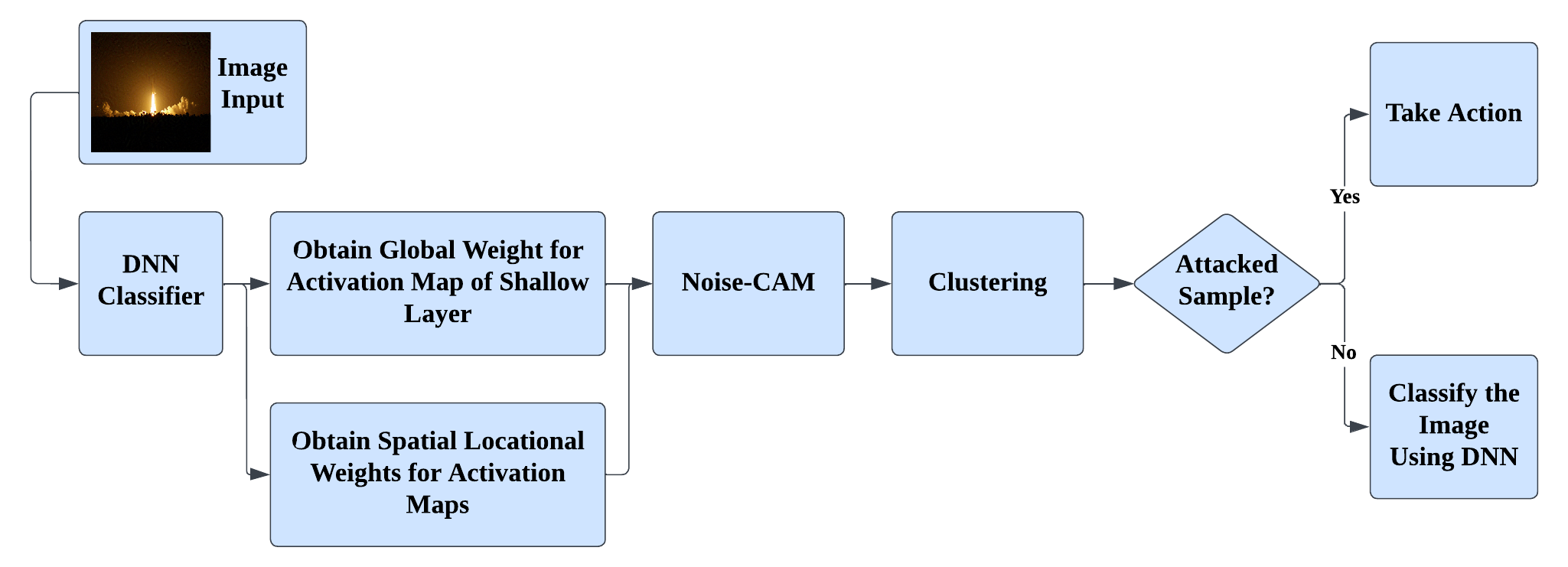}
    \caption{Detecting adversarial examples with NoiseCAM}
    \label{figAD_WorkFlow}
\end{figure*}

\begin{figure*}[b]
    \centering
    \includegraphics[width = 0.9\textwidth]{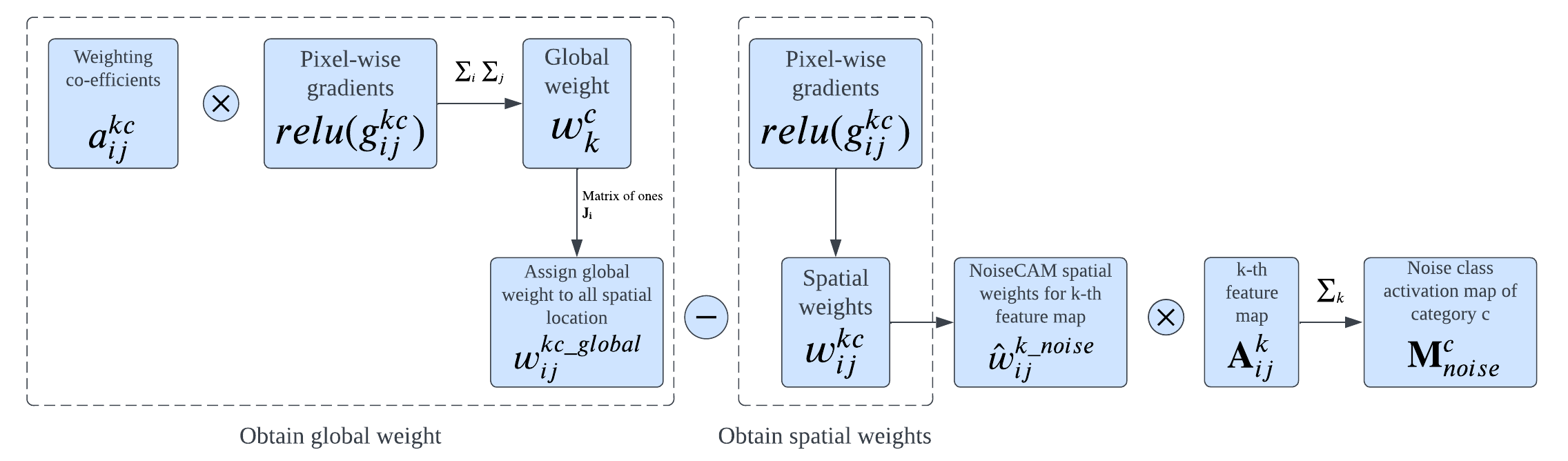}
    \caption{The workflow of NoiseCAM}
    \label{figNoiseCAMworkflow}
\end{figure*}

Our observations revealed that both Gaussian noise and adversarial perturbations can cause behavior deviations, while Gaussian noise does not easily cause a wrong classification result. For each layer, we used the median value of its degree of behavior deviation under Gaussian noise as the reference threshold to quantify whether it is compromised, i.e., whether its behavior deviates more severely than when it is under the same amount and strength of Gaussian noise. 

We also employ behavior deviation to detect adversarial examples. Specific steps are as follows:
\begin{enumerate}[\textbf{Step} 1:]
    \item \textit{Image cleaning: }Given an input image, $\mathbf{I_o}$, we generated a clean version $\mathbf{I_c}$ of it using PCA to compress and retain 99\% of its original information.
    \item \textit{Noise extraction: }We extract the noise component from the original input as $\mathbf{I_n} = \mathbf{I_o} -\mathbf{I_c}$.
    \item \textit{Generating benign noisy samples: }We derive the statistical properties of $\mathbf{I_n}$ as in Equation (\ref{eqGaussianNoise}) and generate statistically similar Gaussian random noise $\mathbf{N_g}$ and add it back to $\mathbf{I_c}$. In this way, we generate a benign noisy image. The process is repeated at least 50 times to obtain sufficient samples.
    \item \textit{Obtaining input-specific statistical model: }We let benign noisy samples and $\mathbf{I_c}$ pass through the model and obtain the statistical properties of the behavior deviation of the most sensitive (vulnerable) convolution layer. In the VGG-16 model, the layer selected as a probe is the first convolution layer in block 5 according to our experiments in Section \ref{sectEvaBehvDev}.
    \item \textit{Comparing: }We feed the original image into the model. If the original image compromises the last convolution layer by causing sufficient drift. We then decide that this input is an adversarial example.
\end{enumerate}

\subsection{NoiseCAM for Adversarial Example Detection}
\label{sectNoiseCAM}


Along with behavior deviation modeling, we present an XAI-based algorithm, called NoiseCAM, for detecting adversarial examples as shown in Fig.~\ref{figAD_WorkFlow}. Our proposed workflow capitalizes on gradient information from the DNN classifier's internal layers. To extract necessary information, we combine GradCAM++ \cite{chattopadhay2018grad} and LayerCAM \cite{9462463} as follows: 
\begin{enumerate}
    \item \textit{Globally Weighted CAM:}
We let ${y^c}$ be the prediction score of the DNN classifier of the target category ${c}$, and for a selected convolution layer, ${\mathbf{A}_k}$ is the ${k}$-th output feature map of the layer output tensor. The spatial gradient ${g^{kc}_{ij}}$ with respect to ${\mathbf{A}_k}$ is as:
\begin{align}
     \label{eqGradient}
     g^{kc}_{ij} = \dfrac{\partial y^c}{\partial \mathbf{A}^k_{ij}}
\end{align}
where $g^{kc}_{ij}$ is the partial derivative of the prediction score of category ${c}$ with respect to the pixels of the ${k}$-th feature map of the selected convolution layer. In the NoiseCAM algorithm, we use the category with the highest prediction score to generate $g^{kc}_{ij}$. We further process this spatial gradient tensor using similar method as in GradCAM++\cite{chattopadhay2018grad}. Specifically, we first compute the enhanced spatial gradients as:. 
\begin{align}
     \label{eqCo-efficient}
     a^{kc}_{ij} = \frac{(g_{ij}^{kc})^2}{2(g_{ij}^{kc})^2 + \sum_a \sum_b \mathbf{A_{ab}^{k}}(g_{ij}^{kc})^3}
\end{align}
where both ($a$, $b$) and ($i$, $j$) are the spatial location of the feature map. We then compute the weights of each channel in the output tensor ${w^c_k}$ as in Equation (\ref{eqCo-efficient}). The final class activation map that captures activations positively contribute to the classification score $Y^C$ is obtained by a linear aggregation along all ${k}$ feature maps as in Equation (\ref{eqGrad-CAM++Final})
\begin{align}
     \label{eqGrad-CAM++Weight}
     w^c_k = \sum_i \sum_j a^{kc}_{ij} \cdot relu(g^{kc}_{ij})\\
     \label{eqGrad-CAM++Final}
     M^c = \mathbf{ReLU}(\sum_k w^c_k \cdot \mathbf{A}^{k})
\end{align}
According to our experiments, adversarial perturbations severely affect feature maps in shallow layers. The global weights used in Grad-CAM++ algorithm will treat all activated pixels as equal. In our experiments, the Grad-CAM++ algorithm generally outperforms Grad-CAM in deriving $w^c_k$ when multiple objects are in the input.

    \item \textit{Pixel-wisely weighted CAM:} This CAM uses the same mathematical process as in LayerCAM \cite{9462463}. Each feature map ${\mathbf{A}^k_{ij}}$ is first weighted pixel-wisely as:
\begin{align}
     \label{eqLayer-CAM}
     w^{kc}_{ij} = relu(g^{kc}_{ij})\\
     \label{eqLayer-CAM2}
     \mathbf{\hat{A}^k_{ij}}= w^{kc}_{ij} \cdot \mathbf{A^k_{ij}}     
\end{align}
where ${w^{kc}_{ij}}$ is the pixel-wise spatial weight matrix that has the same size with the $k$-th feature map with respect to $y^C$.  LayerCAM \cite{9462463} showed that the spatial weight method could eliminate most of the noise and preserve fine-grained information. 

\end{enumerate}

Our proposed NoiseCAM combines the information from globally and pixel-wisely weighted CAMs. The brief workflow of NoiseCAM is given in Fig~\ref{figNoiseCAMworkflow}. We first obtain the global weight of the ${k}$-th activation map from shallow layers. Then, we obtain the pixel-wise spatial weight of the ${k}$-th activation map from the same layer. To obtain the noise weight ${\hat{w}_{ij}^{k\_noise}}$, we first assign the global weight to all spatial locations (i, j) on the activation map, which is formulated as:
\begin{align}
     \label{eqGlobalWtoSpatialW}
     w_{ij}^{kc\_global}= w^c_k \cdot \mathbf{\Omega}^k
\end{align}
where $\mathbf{\Omega}^k$ is an all-one matrix with the same size as the $k$-th feature map. Then we subtract ${w_{ij}^{kc \_global}}$ with the pixel-wise spatial weight ${relu(g^{kc}_{ij})}$ and form a spatial noise weight matrix $\hat{w}_{ij}^{k\_noise}$ as:
\begin{align}
     \label{eqNoiseCAMW}
     \notag \hat{w}_{ij}^{k\_noise} &= w_{ij}^{kc\_global} - w^{kc}_{ij}\\
     &=  w_{ij}^{kc\_global} - relu(g^{kc}_{ij})
\end{align}
where $w^{kc}_{ij}$ is defined in Equation (\ref{eqLayer-CAM}). A linear summation can be performed to obtain the final noise activation map with respect to the target category $c$.
\begin{align}
     \label{eqNoiseCAMFinal}
    \mathbf{M_{noise}^c}=\mathbf{ReLU}(\sum_k \hat{w}_{ij}^{k \_noise} \cdot \mathbf{A}_{ij}^{k})
\end{align}

\begin{figure}[]
    \centering    \includegraphics[width=0.85\linewidth]{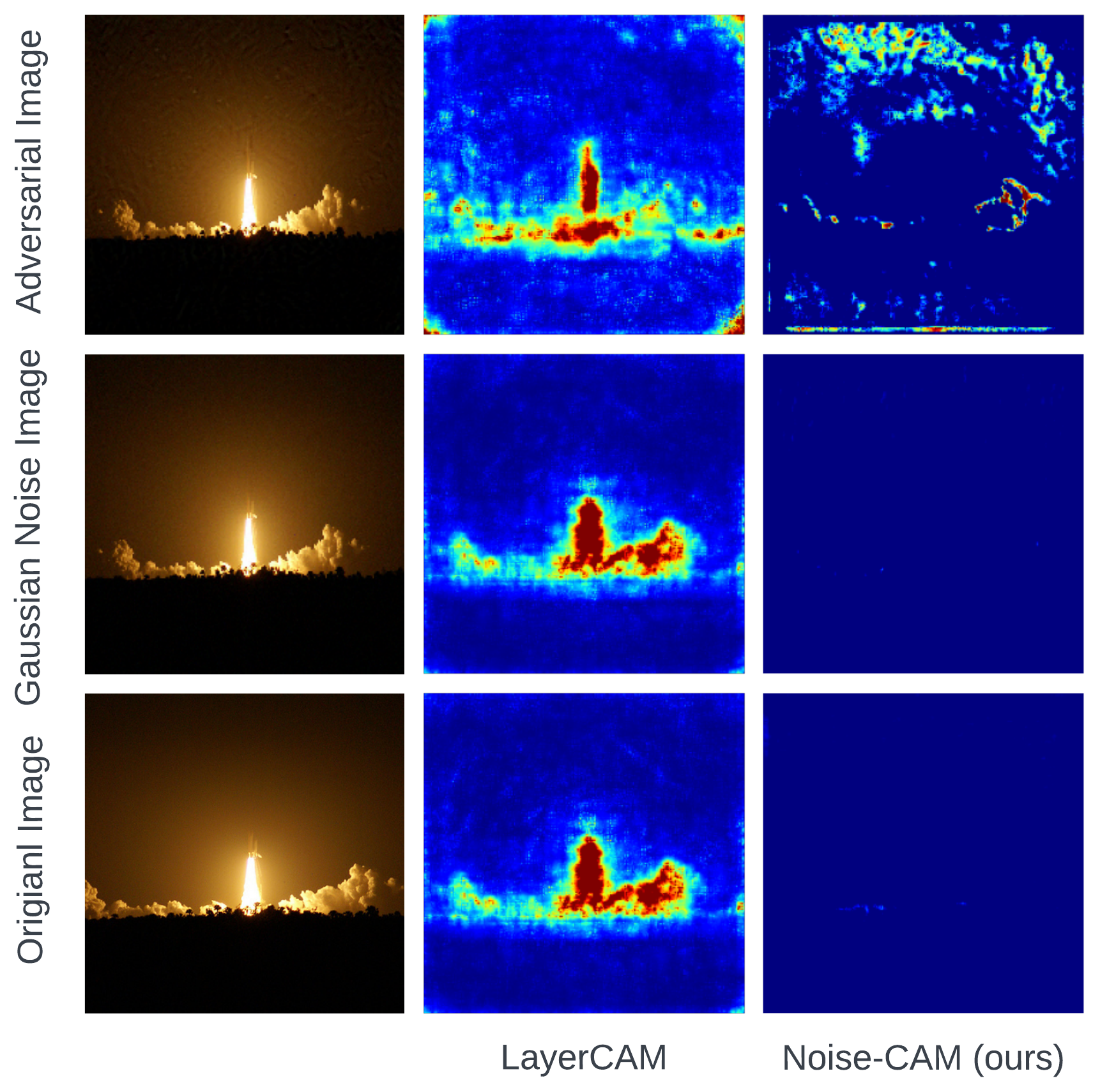}
    \caption{Comparison of adversarial perturbation detection on NoiseCAM and GradCAM++. Top row: the adversarial example (prediction: lampshade) and its class activation map and NoiseCAM map. Center row: original image mixed with Gaussian noise (prediction: space shuttle) and corresponding maps. Bottom row: original image (prediction: space shuttle) and corresponding maps. }
    \label{figDiffInput}
\end{figure}

Mathematically, for an input image with adversarial noise, a globally weighted CAM contains both adversarial perturbations and fine-grained class activation details, while a pixel-wisely weighted CAM only contains the fine-grain details without perturbation noise. Consequently, the subtraction of the two CAMs could expose adversarial perturbations as in Figure~\ref{figDiffInput}. As depicted, NoiseCAM exposes and highlights adversarial noise patterns that do not belong to the subject of interest. On the other hand, NoiseCAM generates a blank map for input with Gaussian noise and original image input.

Finally, the noise activation map, $\mathbf{M_{noise}^c}$, is processed by the DBSCAN \cite{schubert2017dbscan} clustering algorithm, we extract the number of effective noise clusters from the noise activation map, if the number of effective noise cluster exceeds 3, we then judge that the input is an adversarial example. Specifically, we set the scanning radius to 2 and the number of neighboring points to 3 as well.

\section{Evaluation}
\label{sectEED}
This section examines the behavior deviation of the VGG-16 classifier when subjected to varying degrees of adversarial attacks and Gaussian noise. The goal is to determine the boundary between adversarial and natural noise. Moreover, we compare the efficacy of NoiseCAM and behavior deviation modeling in detecting adversarial examples.

\subsection{Seed Selection}
The ImageNet2012 validation dataset \cite{imagenet75:online} contains 5,000 images in 1,000 categories. We use all 5,000 images to derive their adversarial examples. Unfortunately, not all images can find their corresponding adversarial examples within the pre-defined magnitude of perturbation. Ultimately, we derived at least one adversarial example from 48\% of the dataset.

\subsection{Behavior Deviation Under Different Attacks}
\label{sectEvaBehvDev}
Using the methods in Section \ref{sectBDD}, we found that Gaussian noise and adversarial perturbations can cause deviations in the VGG-16 model's behavior. Adversarial perturbations can further drift the network behavior and play a vital role in misleading the classifier. 

\begin{figure}[]
\centering  
\subfloat[]
{%
    \includegraphics[width=0.8\linewidth]{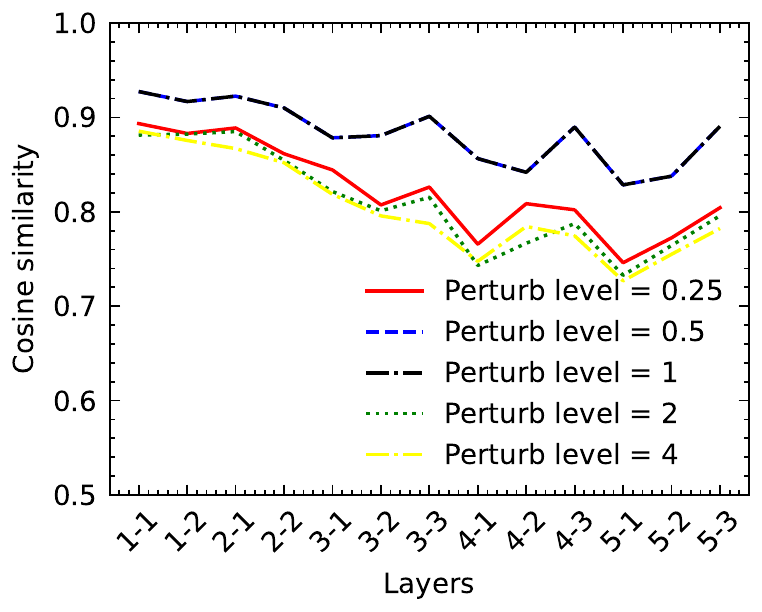}
    \label{figLineplotNormal}
}\\
\subfloat[]
{
    \includegraphics[width=0.8\linewidth]{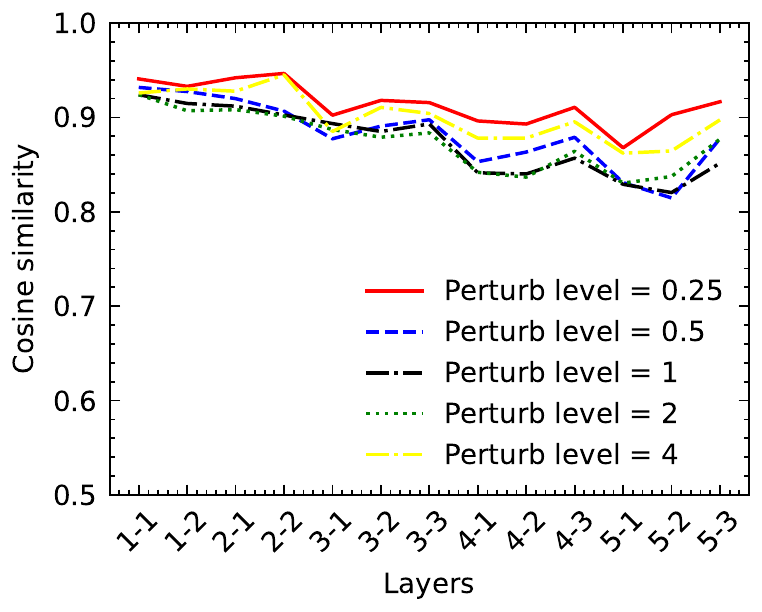}
    \label{figLineplotRandom}
}%
\caption{Comparison of behavior deviation on: (a) adversarial examples and (b) images with Gaussian noise with different attack strength (x\text{-}axis 1\text{-}1 represents the Block1\_Conv1 layer in VGG-16).}
\label{figCosSimLineplot}
\end{figure}

According to Figure~\ref{figLineplotNormal} and \ref{figLineplotRandom}, the comparison of behavior deviation indicates that adversarial perturbation causes more severe behavior deviations than statistically similar Gaussian noise. Although it is generally believed that a stronger adversarial attack strength can cause the behavior of the neural network to deviate further, our experiments show that the behavior deviation at perturbation level 0.25 is even more significant than at levels 2 and 4. Gaussian noise can cause a deviation in the behavior of the neural network, but this deviation is not significant to lead to a classification error. 

Further behavior deviations can be observed in deeper layers of the network, as shown in Figures~\ref{figLineplotNormal} and \ref{figLineplotRandom}. However, we do notice some fluctuations, which indicates that there are some specific layers that are more vulnerable and easier to compromise. For example, the first convolution layer in the fifth block is sensitive to adversarial perturbations and Gaussian noise simultaneously.


\begin{figure}[]
    \centering
    \includegraphics[width=0.8\linewidth]{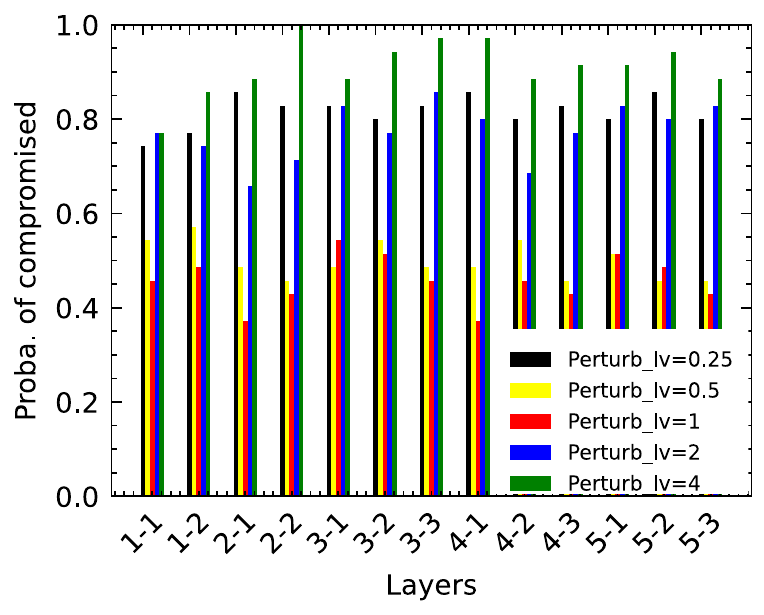}
    \caption{Compromise probability of convolution layers in the VGG-16 network.}
    \label{figVulnerBarplot}
\end{figure}


We also discover that the VGG-16 network can be misled by compromising only a few intermediate layers. We analyze the probability of compromise on each network layer and derive Figure~\ref{figVulnerBarplot}. As depicted, the adversarial examples with the lowest attack strength have approximately 40\% of the chance to compromise any layer. This probability increases significantly and reaches 80\% when we increase the attack strength to be greater than 2 or less than 0.25.



\subsection{Adversarial example detection}

\begin{figure}[]
\centering  
\subfloat[]
{%
    \includegraphics[width=0.9\linewidth]{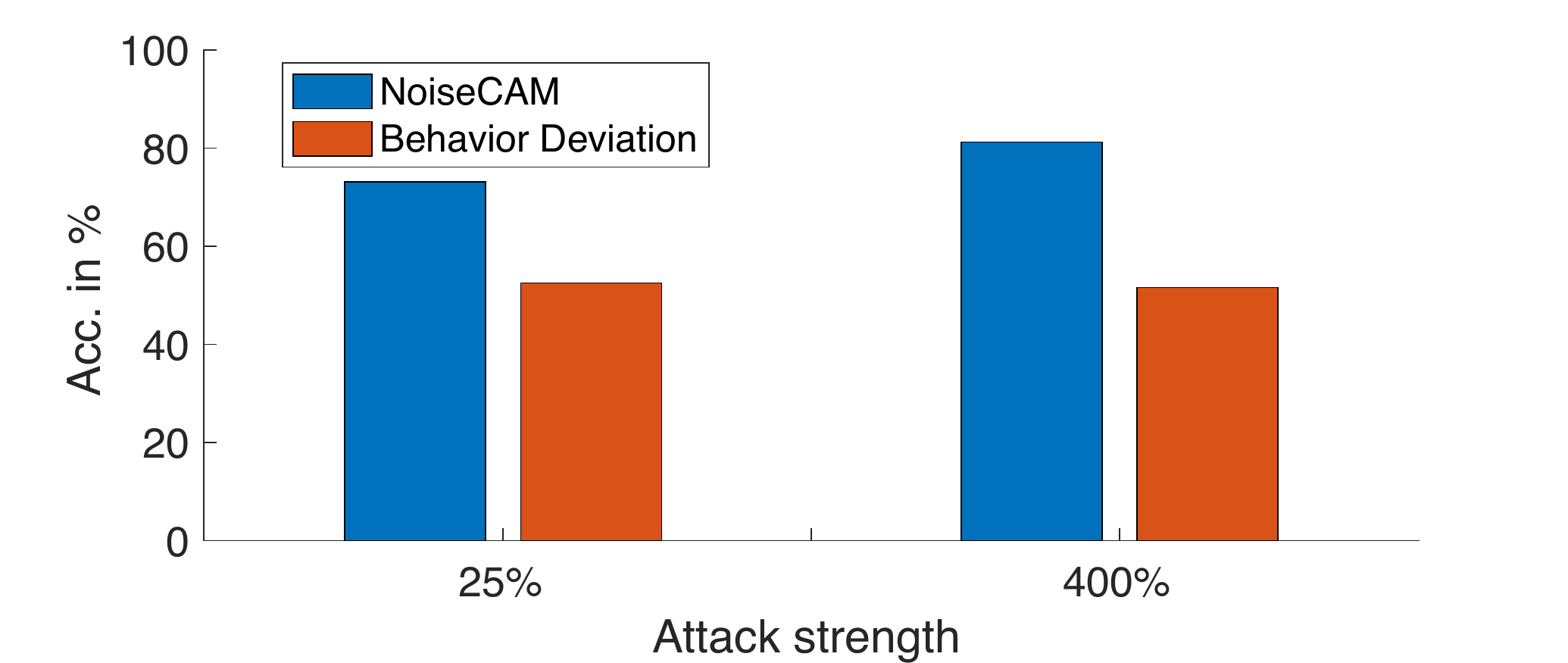}
    \label{figNoiseCAMAcc}
}\\
\subfloat[]
{
    \includegraphics[width=0.9\linewidth]{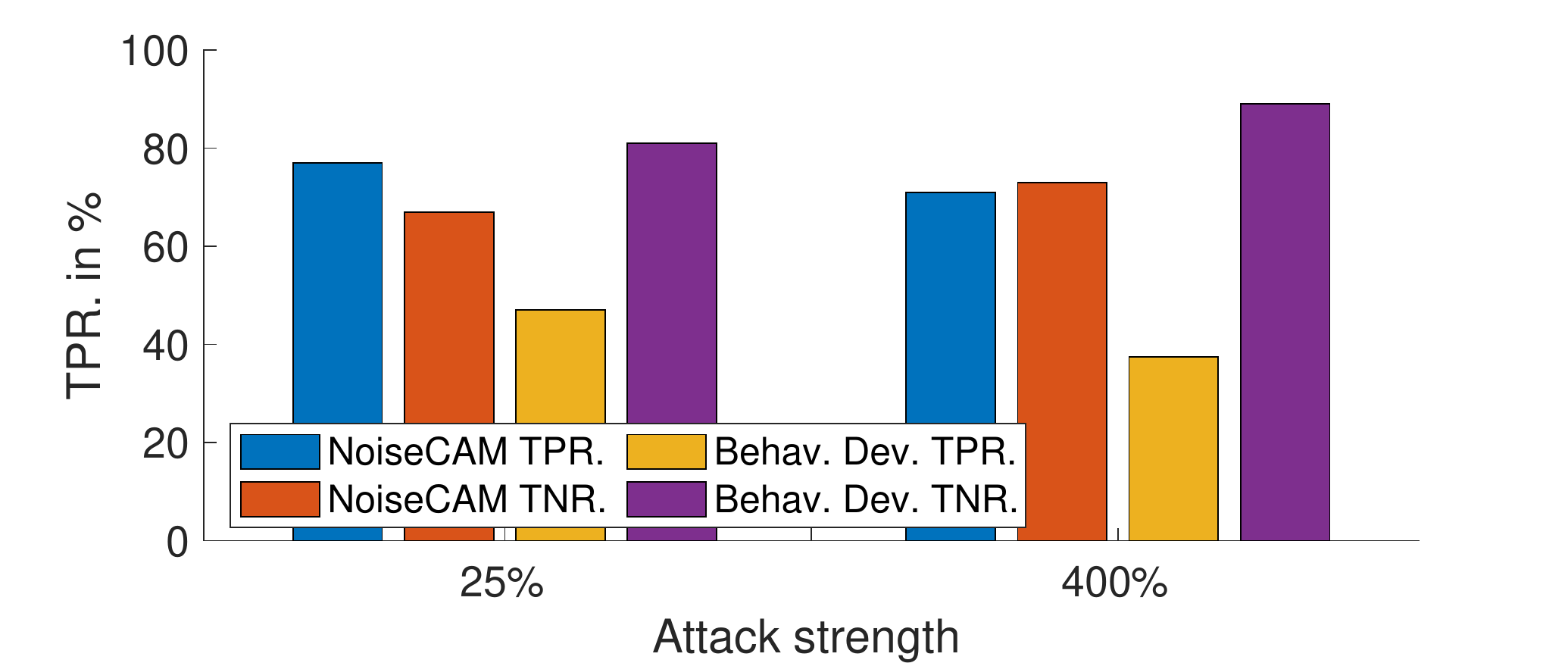}
    \label{figNoiseCAMTPTNR}
}%
    \caption{Performance comparison of adversarial example detection}
    \label{figNoiseCAMEvalAll}
\end{figure}

\begin{figure*}[]
    \centering
    \includegraphics[width = 0.85\textwidth]{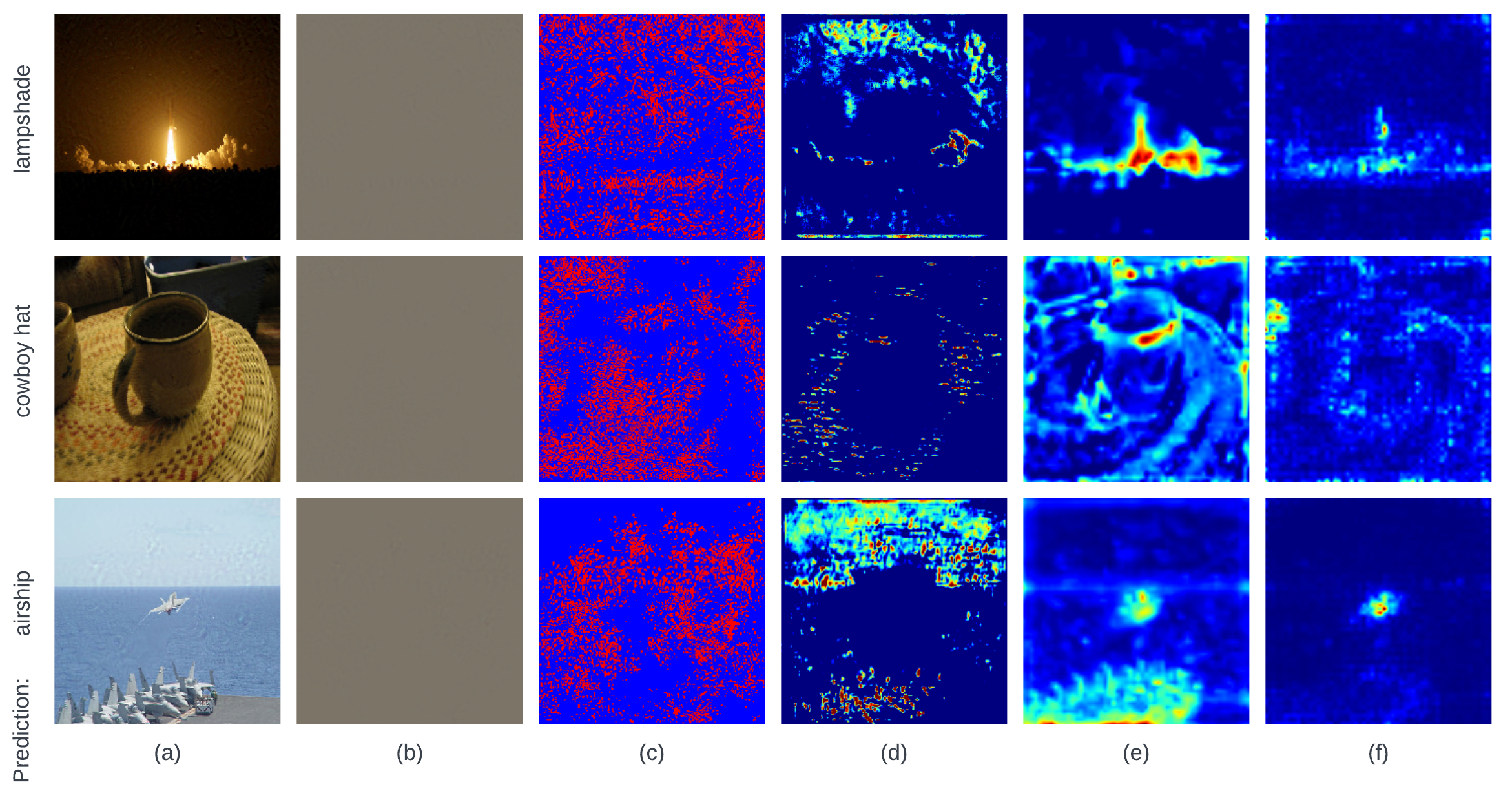}
    \caption{The class activation map on adversarial perturbations. (a) Adversarial images. (b) Adversarial perturbation. (c) Adversarial perturbation visualization. (d) Activation map generated by NoiseCAM. (e-f) Class activation maps from the third convolution layer of VGG-16 model generated by GradCAM++ and LayerCAM.}
    \label{figPerturb_shape}
\end{figure*}

We evaluated the adversarial example detection performance using an augmented dataset consisting of our seed images and their modifications. This test set comprises seed images mixed with adversarial perturbations and Gaussian noise of varying intensities. Our approach involves feeding the input image into a pre-trained VGG-16 model on the ImageNet dataset, applying NoiseCAM on the VGG-16 network's third convolution layer, and using the behavior deviation modeling method on the same test set but on the most vulnerable convolution layer (Block5\_1). Figure~\ref{figPerturb_shape} showcases examples of adversarial example detection with intermediate results. By combining the responses of GradCAM++ and LayerCAM, adversarial perturbations are effectively highlighted. Figure~\ref{figNoiseCAMEvalAll} provides a brief comparison of the two methods, demonstrating that NoiseCAM has a higher detection accuracy than behavior deviation analysis, as shown in Figure~\ref{figNoiseCAMAcc}. Notably, the behavior deviation modeling method has a lower true positive rate and a higher true negative rate than NoiseCAM, as seen in Figure~\ref{figNoiseCAMEvalAll}. Interestingly, increasing the attack strength from 25\% to 400\% slightly reduces the true positive rate of the behavior deviation modeling method, but has no significant impact on the performance of the NoiseCAM approach.

\section{Conclusion and Future Work}

In this work, we model the behavior of a VGG-16 model using an explainable AI approach when its input is mixed with adversarial perturbations from white-box attacks and statistically similar Gaussian noise. We proposed two approaches, behavior deviation modeling and NoiseCAM to detect adversarial examples and prevent the model from being misled. NoiseCAM can highlight adversarial perturbations mixed in the input while it's not sensitive to random noise. We found that NoiseCAM is more reliable than behavior deviation modeling.  Going forward, we aim to apply our approach as a vulnerability detection and securing tool for neural networks across various models.


\label{sectCC}

\section*{Acknowledgment}

This research was partially supported by the National Science Foundation under Grant No. 2309760.

\bibliographystyle{IEEEtran}
\bibliography{IPCCC2022.bib}

\end{document}